# Safe, Efficient, Comfort, and Energy-saving Automated Driving through Roundabout Based on Deep Reinforcement Learning

Henan Yuan, Penghui Li, Bart van Arem, *Senior Member, IEEE*, Liujiang Kang, and Yongqi Dong

*Abstract*—**Traffic scenarios in roundabouts pose substantial complexity for automated driving. Manually mapping all possible scenarios into a state space is labor-intensive and challenging. Deep reinforcement learning (DRL) with its ability to learn from interacting with the environment emerges as a promising solution for training such automated driving models. This study explores, employs, and implements various DRL algorithms, namely Deep Deterministic Policy Gradient (DDPG), Proximal Policy Optimization (PPO), and Trust Region Policy Optimization (TRPO) to instruct automated vehicles' driving through roundabouts. The driving state space, action space, and reward function are designed. The reward function considers safety, efficiency, comfort, and energy consumption to align with real-world requirements. All three tested DRL algorithms succeed in enabling automated vehicles to drive through the roundabout. To holistically evaluate the performance of these algorithms, this study establishes an evaluation methodology considering multiple indicators such as safety, efficiency, and comfort level. A method employing the Analytic Hierarchy Process is also developed to weigh these evaluation indicators. Experimental results on various testing scenarios reveal that the TRPO algorithm outperforms DDPG and PPO in terms of safety and efficiency, and PPO performs best in terms of comfort level. Lastly, to verify the model's adaptability and robustness regarding other driving scenarios, this study also deploys the model trained by TRPO to a range of different testing scenarios, e.g., highway driving and merging. Experimental results demonstrate that the TRPO model trained on only roundabout driving scenarios exhibits a certain degree of proficiency in highway driving and merging scenarios. This study provides a foundation for the application of automated driving with DRL in real traffic environments.**

## I. Introduction

Automated vehicles (AVs) promise to mitigate a myriad of uncontrollable factors associated with human operations, including human error and subjective judgment. The technology underpinning automated driving constitutes an amalgamation of multiple disciplines, with the system primarily composed of perception, planning, decision-making, and control modules. The decision-making module, governing actions such as throttle and braking control, vehicle steering, and signal light operation, is particularly critical. Its task is not only to define the driving trajectory but also to respond to unexpected scenarios, making it the key element.

Deep reinforcement learning (DRL), an intersection of deep learning's capabilities of capturing features and reinforcement learning's decision-making aptitude, has been widely acclaimed in the field of automated driving. It has been witnessed that DRL even outperformed human decision-making in numerous applications. A typical example would be AlphaGo [1], the first artificial intelligence to defeat a human professional Go player, employing a DRL algorithm. Through six million rounds of learning and environmental interaction, AlphaGo honed its capability to triumph over world champions.

The existing studies in the domain of automated driving with DRL have broadly addressed various control tasks and driving scenarios.

### A. DRL for Different Driving Tasks\

DRL has been deployed in a variety of control tasks regarding driving. Sallab et al. [2] employed DRL to investigate lane-keeping tasks, utilizing Deep Q-Networks (DQN) for discrete action control and the Deep Deterministic Actor-Critic (DDAC) approach for continuous actions. Wang et al. [3] delved into lane-changing tasks, highlighting the capability of DRL to manage anomalous scenarios. The same research group also integrated Long Short-Term Memory (LSTM) with DQN to tackle ramp merging [4]. Their architecture accounts for the influence of interactive environments on long-term rewards to establish the optimal policy.

Ngai and Yung [5] utilized DRL to train AVs for overtaking maneuvers. Their findings suggest that Q-learning enables the agent to make judicious decisions, preventing collisions with surrounding objects. Moreover, the agent can complete overtaking within the stipulated time, maintaining a stable heading angle during the process.

Moreira [6] conducted tests on several DRL algorithms, e.g., Soft Actor-Critic (SAC), Deep Deterministic Policy Gradient (DDPG), and Twin Delay Deep Deterministic Policy Gradient (TD3), for automated parking. The proposed reward function was determined by the angle between the agent's driving direction and the correct direction. Results indicate that the TD3 algorithm, with its rapid convergence rate, is most suited to the automated parking scenario.

### B. DRL for Various Driving Scenarios

Diverse automated driving scenarios have been studied using DRL. Fayjie et al. [7] utilized DQN for

*Research supported by Applied and Technical Sciences (TTW), a subdomain of the Dutch Institute for Scientific Research (NWO), Grant/Award Number: 17187.

Henan Yuan is with the School of Traffic and Transportation, Beijing Jiaotong University, Beijing, 100044 China (e-mail: 19252022@bjtu.edu.cn).

Penghui Li is with the School of Traffic and Transportation, Beijing Jiaotong University, Beijing, 100044 China (e-mail: penghuil@bjtu.edu.cn).

Bart van Arem is with the Department of Transport and Planning, Delft University of Technology, Delft, 2628 CN the Netherlands (e-mail: b.vanarem@tudelft.nl).

Liujiang Kang is with the School of Traffic and Transportation, Beijing Jiaotong University, Beijing, 100044 China (e-mail: ljkang@bjtu.edu.cn).

Yongqi Dong is with the Department of Transport and Planning, Delft University of Technology, Delft, 2628 CN the Netherlands (corresponding author, phone: +31 616619399; e-mail: y.dong-4@tudelft.nl).

decision-making to train car driving in urban environments. They used *Unity* to design a city-like structure with buildings, trees, and street lights, and utilized lidar and camera data as the state space validating neural networks' effectiveness in such settings. Konstantinos et al. [8] examined automated driving in highway scenarios. Tram et al. [9] applied DQN to predict vehicle trajectories at intersections, highlighting the superior success rate of the Deep Recurrent Q Network.

Kamran et al. [10] and Jiang et al. [11] focused on driving through unsignalized intersections, using DQN and progressive value-expectation estimation multi-agent cooperative control (PVE-MCC), respectively. Chen et al. [12] addressed on-ramp merging with a multi-agent DRL, considering mixed traffic conditions of human-driven and automated vehicles running together.

### C. *Roundabout Driving*

When it comes to the roundabout, which is integral to urban traffic infrastructure, Elvik [13] has shown that roundabouts can effectively reduce the probability of serious traffic accidents. However, the intricate interweaving with other road users and exit selection at roundabouts pose significant challenges to automated driving. Given the impracticality of manually recording all possible scenarios in the state space, DRL emerges as a suitable approach for automated driving decision-making through roundabouts. However, very limited studies have tackled roundabouts by DRL, only three were identified, i.e., García et al. [14] used a Q-Learning algorithm, Capasso et al. [15] utilized an Asynchronous Advantage Actor-Critic (A3C)-based vehicle mobility planning module, and Wang et al. [16] proposed SAC algorithm combined with interval prediction and self-attention mechanism for roundabouts driving. There are still noticeable gaps in the complex roundabout driving scenarios, especially when it comes to employing and comparing different DRL algorithms in the context of mixed traffic of human-driven and AVs and considering integrated rewards. Furthermore, the domain adaption possibilities, i.e., evaluating the feasibilities of transferring the algorithms trained in the roundabout driving to other scenarios (e.g., highway driving) remains unexplored.

As a preliminary exploration, this study attempts to tackle these critical research gaps and tries to harness DRL to facilitate the navigation and control of AVs through roundabouts, underpinned by a carefully designed reward function that accounts for the unique challenges presented in this complex traffic scenario. For that, an integrated rewards function considering safety, efficiency, comfort level, and the energy consumption is developed. Three state-of-the-art DRL algorithms, i.e., DDPG, Proximal Policy Optimization (PPO), and Trust Region Policy Optimization (TRPO), are then implemented. Experiments show that TRPO outperforms other DRLs in tackling automated driving through roundabouts and the ablation study also demonstrated the transferability of the developed model in handling other driving scenes.

## II. Methodology

### A. *System Architecture*

This study aims to develop safe, efficient, comfortable, and energy-saving driving models for AVs passing through roundabouts. Different DRL algorithms including DDPG, TRPO, and PPO were implemented and tested using a well-designed rewards function. The overall proposed system architecture is shown in Figure 1.

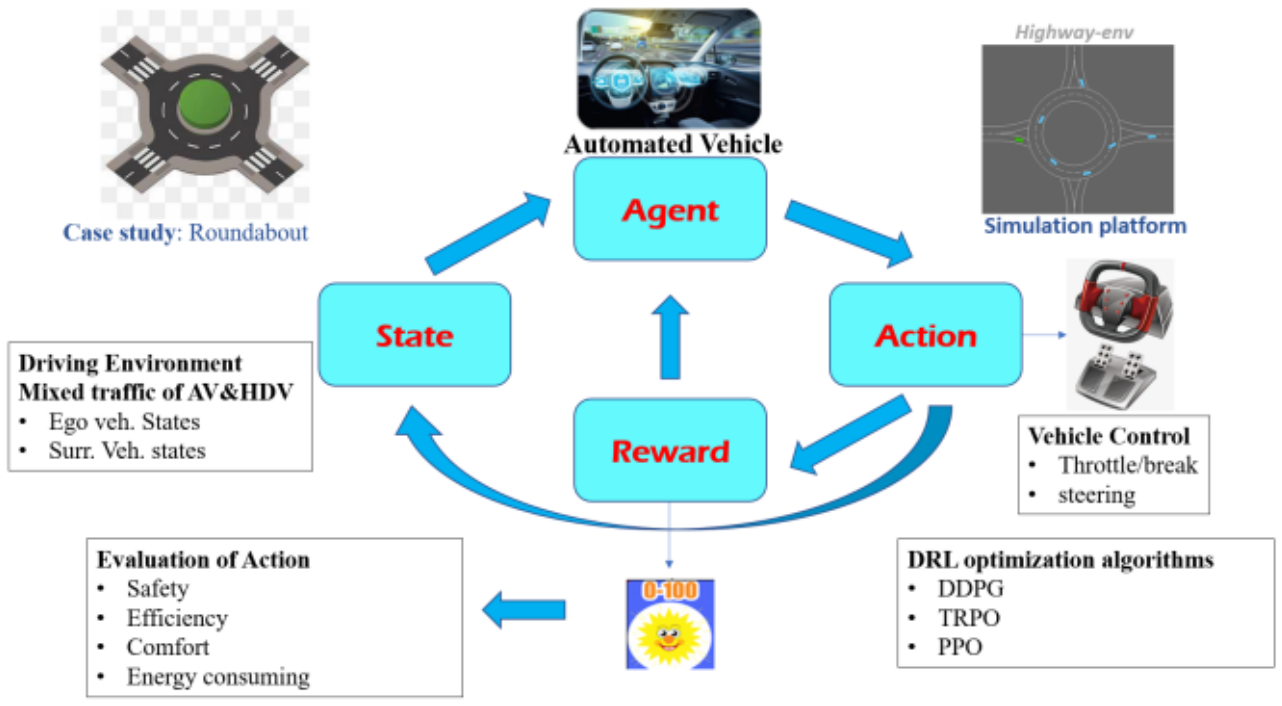

Figure 1. Illustration of the overall system architecture

The deep reinforcement learning was implemented through the PyTorch deep learning framework and the *highway-env* [17] simulation platform. The DRL algorithms are instantiated via the *stable-baselines3* [18] reinforcement learning library. *Highway-env* is a Python library comprising a collection of environments for automated driving, encompassing several typical road scenarios: highway, merge, roundabout, parking, intersection, and racetrack. Predominantly, this study trained and tested the DRL models on the roundabout scenario, while scenarios such as highway merging are used to test and verify the model's versatility.

### B. *DRL*

DRL is a specialized machine learning algorithm designed to aid agents in decision-making. Through interactive training between the agent and its environment, DRL can enhance the agent's decision-making capacities. Specifically, in each training process timestep, the agent performs an action, after which the environment determines the agent's state and provides a reward value. This reward value assesses the agent's state at that timestep. Subsequently, the agent adjusts the policy network's parameters by computing or estimating the cumulative reward value. This process allows the model to maximize the achievable reward, optimize the decision-making strategy, and determine subsequent actions. This study implements, customizes, and compares three DRLs, i.e., DDPG, TRPO, and PPO, regarding roundabout driving.

### C. *Environment, State, and Action Settings*

#### 1) *Environment*

The roundabout environment, a subclass of *AbstractEnv* in the *highway-env* library, simulates a vehicle navigating through roundabouts. It allows customization of road shape, parameters, and vehicle behavior, as well as the reward function and termination conditions of reinforcement learning. The driving task for the trained agent is to achieve safe, quick, and efficient driving while avoiding collisions and adhering to a predetermined route as closely as possible. To train the autonomous vehicle's interaction capabilities with surrounding traffic, several vehicles are randomly added to the roundabout environment. These vehicles, defined by the *highway-env* library, can demonstrate simple driving behavior and collision avoidance in low-density traffic. A schematic diagram of the roundabout with surrounding vehicles is shown in Figure 2.

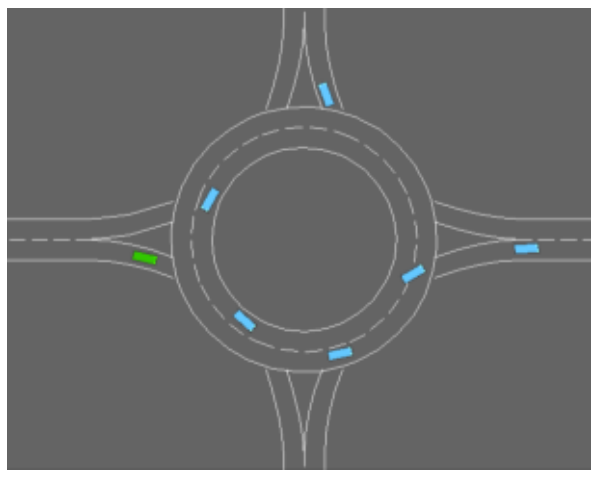

Figure 2. Roundabout with surrounding vehicles

*2) State space*

In DRL, state space encapsulates observable data used by the agent to determine corresponding actions via neural network processing. Regarding automated driving, state space consists of real-time information, such as vehicle speed, acceleration, heading angle, and surrounding traffic conditions. This study adopts seven aspects of features to represent the state spaces shown in Table 1.

TABLE I. STATE SPACE AND FEATURES

| **Feature name** | **Feature meaning** |
|---|---|
| Presence | Indicates the presence or absence of the vehicle |
| *x* | The position of the vehicle on the *x*-axis |
| *y* | The position of the vehicle on the *y*-axis |
| *vx* | The speed of the vehicle on the *x*-axis |
| *vy* | The speed of the vehicle on the *y*-axis |
| Lane_heading_ difference | The difference between the front direction of the car and the direction of the lane |
| Lane_distance | The distance from the vehicle to the lane centerline |

*3) Action space*

The highway-env environment offers three types of action spaces: *Discrete Action*, *Discrete Meta Action*, and *Continuous Action*. This study employs a hybrid approach using both discrete and continuous actions to train distinct driving tasks.

*Discrete Meta Action* discretizes continuous vehicle behavior into meta-behaviors, such as acceleration, deceleration, lane changes, and maintaining speed. Each meta-action, defined by its duration and a sequence of basic actions, facilitates efficient exploration and learning of complex behaviors while preserving action continuity and interpretability.

*Continuous Action* involves throttle and steering controls. The throttle ranges from -1 (maximum brake) to 1 (maximum throttle), and the steering control ranges from -1 (maximum left turn) to 1 (maximum right turn).

In this study, the continuous action space is mainly used as an action space. Its actions need to ensure that the vehicle can drive along the lane on the roundabout, reach the destination exit, and have the ability to avoid other vehicles on the road.

## D. Rewards Function

In this study, a reward function is crafted specifically for autonomous vehicles' navigating roundabouts. The effectiveness of the reward function is determined by evaluating the driving safety, efficiency, comfort, and energy consumption through the analysis of performance indicators.

*1) Safety rewards*

Vehicle safety is paramount in autonomous driving, hence it accounts for substantial weight in the reward function. In the roundabout driving context, safety is primarily influenced by two factors, i.e., lane-center positioning and time-to-collision (TTC). The lane-centering reward, indicated by $R_{LC}$., can be computed as

$$R_{LC} = 1 - \left(\frac{l_{lateral}}{l_{width}/2}\right)^2 \tag{1}$$

where $l_{lateral}$ is the vehicle's offset to the center of the lane, and $l_{width}$ is the lane width.

The TTC reward is computed as

$$R_{TTC} = 1 - \frac{3}{TTC} \tag{2}$$

If the Time-to-Collision (TTC) exceeds 3 seconds, the TTC reward will fall within the range of 0 to 1. A larger TTC results in a reward closer to 1. Conversely, when TTC is less than 3, the reward becomes negative. And in the event of an imminent collision, the TTC reward will approach $-\infty$.

The total safety reward is a weighted sum of the lane center reward and the TTC reward. The TTC reward constitutes 70% of the $R_{safe}$, while the lane center reward makes up the remaining 30%. The total safety reward can be expressed as:

$$R_{safe} = 0.7 \times R_{TTc} + 0.3 \times R_{LC} \tag{3}$$

*2) Efficiency rewards*

The efficiency reward motivates the AV to move forward, avoiding stationary actions. It mainly rewards high speeds within set limits. When the vehicle's speed is less than or equal to the speed limit, the efficiency reward is set to the ratio of the vehicle's current speed to the speed limit as

$$R_{efficient} = \frac{v_{ego}}{v_{limit}} \tag{4}$$

When the vehicle's speed is greater than the speed limit, the reward value decreases as the speed increases.

$$R_{efficient} = 1 - \frac{v_{ego} - v_{limit}}{v_{max} - v_{limit}} \tag{5}$$

where $v_{ego}$ is the current speed, $v_{limit}$ is the speed limit on the road, and $v_{max}$ is the maximum achievable speed value of the vehicle.

*3) Comfort rewards*

Vehicle comfort, a key performance indicator for automated driving, significantly impacts user experience. This study focuses on smooth acceleration, deceleration, and steering. The reward function considers the rate of change in acceleration/braking and steering. Lower rates of change, indicating smoother movements, yield higher rewards, while higher rates of change result in lower rewards. The calculation of the *Comfort* reward value is as follows

$$diff_{throttle} = \frac{d\ throttle_t}{dt} \tag{6}$$

$$diff_{steering} = \frac{d\ a_{steering}}{dt} \tag{7}$$

$$R_{comfort} = 1 - \frac{diff_{throttle} + diff_{steering}}{4} \tag{8}$$

where $diff_{throttle}$ is the rate of change of the throttle or brake, $a_{steering}$ is the input value of the throttle or brake, $diff_{steering}$ is the rate of change of the steering wheel, and $a_{steering}$ is the input value of the steering wheel.

*4) Energy consumption rewards*

Jiménez [19] indicates that Vehicle Specific Power (VSP) can indirectly reflect vehicle energy consumption, demonstrating a roughly linear positive correlation with specific power. Hence, specific power values can be used to approximate energy consumption. Parameters for this model were calibrated by Jiménez [19]. In this study, the slope resistance term is omitted since road slope is not considered.

$$VSP = v \times (1.1a + 0.132) + 0.000302v^3 \tag{9}$$

For the setting of the reward function, this study considers the maximum specific power value of the vehicle and uses it as a standard to normalize the value of the specific power at the current moment to the range from 0 to 1, and thus

$$R_{energy} = 1 - \frac{VSP}{VSP_{max}} \tag{10}$$

*5) Total integrated rewards*

In the roundabout setting, AVs will enter from any of the four entrances with a predefined exit destination. A destination reward is implemented for the agent to learn to navigate towards its objective when performing continuous actions. This reward is Boolean, i.e., it is set to 1 if the vehicle reaches the target exit and 0 otherwise:

$$R_{arrive} = \begin{cases} 1 & \textit{if the } \text{vehicle} \textit{ has reached the target exit} \\ 0 & \textit{else} \end{cases} \tag{11}$$

The total integrated reward function combines the aforementioned sub-reward functions through a weighted sum. Having closely similar weights for all four sub-reward functions would overcomplicate the reward function and hinder satisfactory model training. Emphasis is placed on safety and efficiency by assigning larger weights, as they are critical elements. The total reward function is calculated as

$$R_{total} = 0.6\,R_{safe} + 0.25\,R_{efficient} + 0.1\,R_{comfort} + 0.05\,R_{energy} + R_{arrive} \tag{12}$$

## III. EXPERIMENT

This study implemented DDPG, TRPO, and PPO, trained them on the *highway-env* platform, and evaluated and compared their performances. The model training and testing are conducted on a laptop with a 12th Gen Intel Core i9-12900H CPU and an NVIDIA GeForce RTX 3070 Ti GPU. In the implementation, model fine-tuning and hyperparameter optimization play a vital role in enhancing the performance of reinforcement learning algorithms. Model fine-tuning adjusts the algorithm model's specifics and structure, while hyperparameter optimization involves selecting and adjusting the hyperparameters within the algorithm for improving performance.

Typical techniques for model fine-tuning include neural network structure adjustment, e.g., tweaking the number of layers, neurons, and activation function, to boost the algorithm's efficacy. In this research, all these three DRL algorithms adopt similar network structures. Specifically, both the actor and critic networks of DDPG are designed with two hidden layers, each containing 64 neurons. TRPO and PPO utilize the Multi-Layer Perceptron (MLP) neural network with two hidden layers, each containing 64 neurons.

In reinforcement learning, hyperparameters are parameters that cannot be optimized iteratively during training and need to be set manually beforehand. Hyperparameter tuning involves adjusting these parameters to enhance algorithm performance. This study employs grid search to optimize hyperparameter values, preserving or excluding hyperparameter combinations based on the decrease or increase of the reward function during training.

## IV. RESULTS AND ANALYSIS

This study conducted a thorough quantitative comparison of the DDPG, PPO, and TRPO algorithms. The evaluation considers factors such as convergence speed during training, driving efficiency, comfort, lane deviation, and collision rate of autonomous vehicles. Due to differences in reward function discount factors across algorithms, this study extracted these metrics during the model testing phase rather than directly comparing average reward values from training.

### *A. Comparison of Convergence Speed*

Evaluating the convergence speed of algorithms is crucial in deep reinforcement learning, given the dependency on high-performance computing resources and the time-intensive nature of training. This study compared the training progression of DDPG, TRPO, and PPO algorithms as shown in Figure 3 illustrating the variations in average reward values over time for each algorithm.

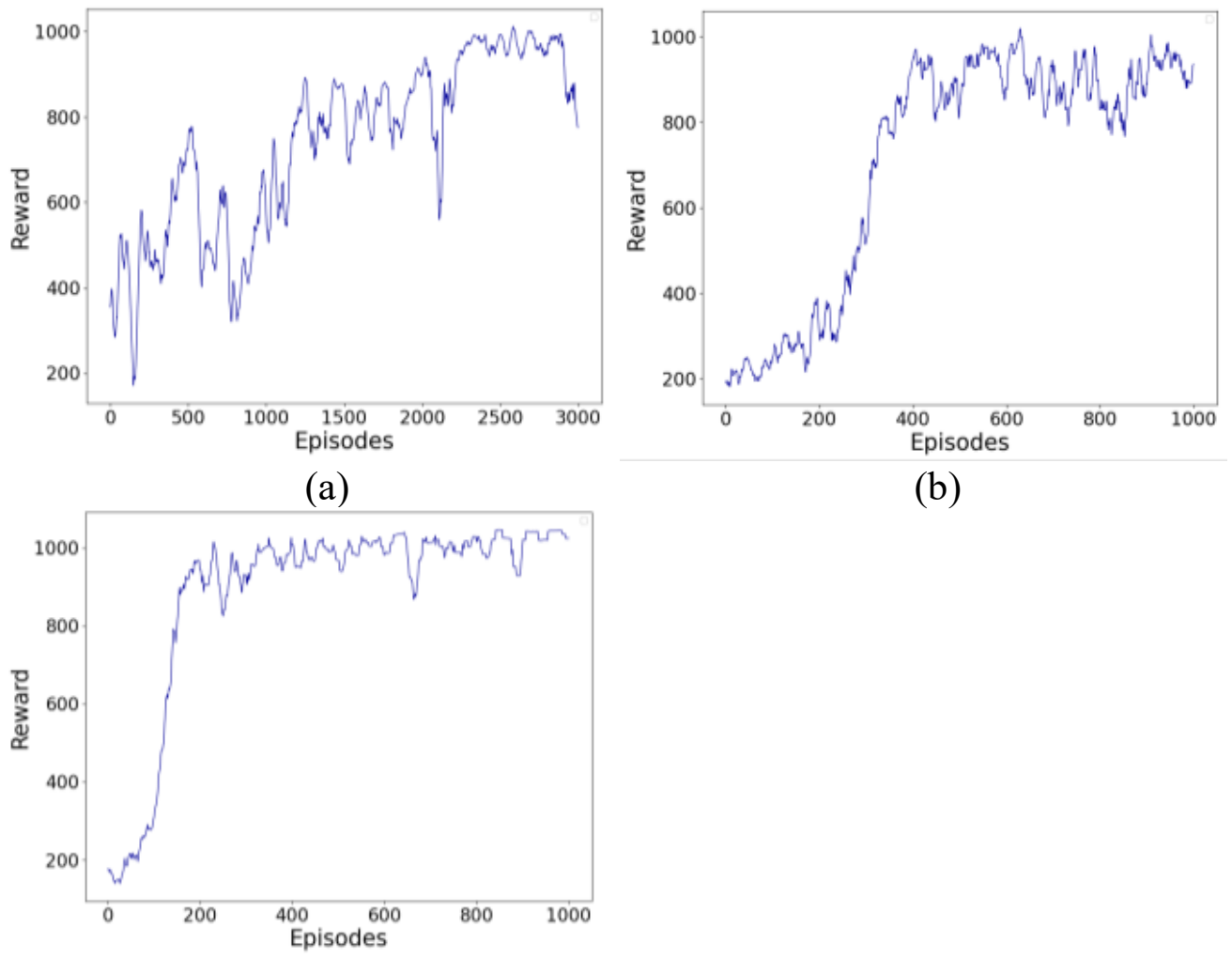

(c)

Figure 3. The training reward value of the three DRL algorithms (a) DDPG, (b) PPO, (c) TRPO

From the observations in Figure 3, all three selected algorithms, i.e., TRPO, PPO, and DDPG, manage to elevate the reward value to approximately 1000 and then maintain a stable range. TRPO, with the quickest convergence speed, reaches a stable maximum reward in about 300 episodes. PPO follows, converging to a similar reward value in approximately 400 episodes. DDPG, with its more parameter-heavy nature, converges slower, stabilizing only after nearly 2300 episodes.

## B. Comparison of Model Performance

The trained model can be invoked for testing using PyTorch's *model.load()* function. The testing records details about the AV's state at each time step, such as throttle and steering states, and collision status. These data enable calculating various performance evaluation metrics like efficiency, safety, and comfort level.

Automated vehicle evaluation typically involves a comprehensive evaluation system that encompasses three stages, i.e., simulation testing, closed road testing, and open lead testing. Each stage requires specific evaluation metrics and weights, along with effective evaluation methods to ensure safety and improve performance.

This study performs simulation testing of automated roundabout driving in the aforementioned environment. In the testing phase, the trained model is invoked. The AVs navigate the roundabout environment based on actions outputted by the invoked model, given the observed state space. For each DRL algorithm, the model is tested over 50 iterations from entering to exiting the roundabout. An observation function extracts the average collision rate, lane-centering loss value, efficiency, comfort, and energy consumption level during these tests. The average values of these five metrics throughout the 50 rounds of testing will be used as the performance indicators. The calculation of the collision rate is shown in Equation (13),

$$Collision\ Rate = 1 - \frac{num_{collision}}{T} \times 10^3 \quad (13)$$

where $num_{collision}$ is the number of vehicle collisions during the entire simulating test, $T$ is the total simulation time step of the 50 rounds of testing (larger than 5000). This calculation converts the collision performance into a score of 0 to 1.

The impact of the above five evaluation indicators on automated driving is different, and thus the weight of each indicator needs to be further analyzed. For that, this study utilized the Analytic Hierarchy Process (AHP) method to determine the weights of the five testing indicators. Details of the AHP process are provided in the supplementary materials. The final estimated weight values are shown in TABLE II.

TABLE II. ESTIMATED WEIGHT VALUES

| Indicator | Weight value |
|---|---|
| Average collision rate test value | 0.4764 |
| Average lane-centering loss | 0.2853 |
| Average efficiency | 0.1428 |
| Average comfort level | 0.0634 |
| Average energy consumption level | 0.0320 |

TABLE III . MODEL TESTING RESULTS

| Indicator / Algorithm | Collision rate | Lane-centering | Efficiency | Comfort | Energy consumption | Total test score |
|---|---|---|---|---|---|---|
| DDPG | 0.43 | 0.8653 | 0.8872 | 0.8846 | 0.8058 | 0.6606 |
| PPO | 0.68 | 0.8385 | 0.8784 | **0.9836** | **0.8103** | 0.7769 |
| TRPO | **0.73** | **0.9322** | **0.9295** | 0.8627 | 0.7995 | **0.8267** |

Each test metric is computed as its average value across all 50 rounds of testing, normalized to a range between 0 and 1. TABLE III shows the testing results of the three selected DRL algorithms.

The results show that TRPO outperforms the other two DRLs in collision rate, lane-centering loss, and efficiency metrics, though it lags slightly in comfort level and energy consumption compared to the other two algorithms. Overall, TRPO achieved the highest integrated test score, surpassing both DDPG and PPO.

DDPG, while defective in terms of collision rate, demonstrates better lane-centering and efficiency performance than PPO, yet falls behind TRPO. While PPO excels in comfort and energy consumption, it lags behind TRPO in terms of the other three metrics. Despite individual algorithm strengths in certain aspects, overall, TRPO performs the best.

For model characteristics and their verification, DDPG uses a deep Q-network to estimate the optimal action-value function, differing from TRPO and PPO, which utilize natural policy gradient algorithms with distinct optimization constraints. For exploration, DDPG applies noise-induced action perturbations suitable for continuous action spaces, although possibly resulting in slower convergence. In contrast, TRPO and PPO use stochastic policies, usually providing more effective global optimal solutions. Unlike DDPG's instability due to hyperparameter sensitivity, TRPO and PPO exhibit robustness and stability thanks to their conservative optimization strategies.

To sum up, TRPO excels in collision rate, lane-centering, and efficiency, and delivers the best overall testing score; PPO is distinct in comfort and energy consumption, and follows TRPO regarding the overall testing score; while DDPG may be hampered by its sensitivity to hyperparameters and less effective exploration strategies leading to the worst overall testing performance.

## C. Ablation Study: Model Adaptability in Other Scenarios

To test the adaptability of the trained TRPO model across other driving scenarios, it was deployed and tested on *highway driving* and *merging* maneuvers on *highway-env*. The TRPO model only trained on roundabout scenarios, showed a certain degree of competence in these new driving tasks. Subjective evaluation by ten experts was done to rate the model's performance across three dimensions: lane keeping, car following, and lane changing (scored 1-3, with 3 being the best). Average scoring results are presented in TABLE IV showing the model's proficient lane-keeping and car-following capabilities in the highway driving scenario. And regarding lane-changing tasks, the model did not perform well. It is understandable, as compared with high driving and merging, there are merely and different lane changes in the training of roundabout driving.

TABLE IV. SUBJECTIVE EVALUATION RESULTS ON MODEL ADAPTABILITY

| Scenario | Lane-keeping | Car-following | Lane-changing |
|---|---|---|---|
| Highway driving | 3.0 | 3.0 | 1.0 |
| Merging | 3.0 | - | 1.5 |

Limited by space, the details of the ablation study are elaborated in the supplementary materials available at https://drive.google.com/drive/folders/1LjalsmioirfXXjoEbXYg-Il4KRx14HMr. In this shared folder, demo videos are also provided to better visualize the results.

## V. Conclusion

This study presents a deep reinforcement learning (DRL) based framework for automated driving through complex roundabout scenarios with surrounding human driven vehicles. Based on the *highway-env* platform, this study designed the corresponding state and action space, together with an integrated multi-factor reward function considering safety, efficiency, comfort, and energy consumption. Using *stable-baselines3*, the study customized and implemented three DRLs, i.e., DDPG, PPO, and TRPO, to achieve automated driving through roundabouts. The models were trained with simulation and fine-tuned by hyperparameter optimization using a grid search approach.

To verify the model performance, this study constructed an evaluation methodology considering different indicators, e.g., safety (collision rate and lane-centering loss), efficiency, comfort level, and energy consumption. Testing results demonstrated that the implemented DDPG, PPO, and TRPO models could all tackle roundabout driving, while, particularly, PPO performed well in terms of comfort level and energy consumption, while TRPO excelled in terms of safety and efficiency, and performed the best in terms of the integrated overall testing score.

To gauge the model's robustness across different driving scenarios, this study tested the TRPO model trained only on roundabout driving in other various driving tasks. The model maintained good performance in highway driving and merging scenarios, albeit not as remarkable as in the roundabout context. With these findings, this paper provides preliminary evidence for developing automated driving with DRL in complex and real traffic environments.